\pdfoutput=1

\documentclass[11pt]{article}

\usepackage{acl}

\usepackage{times}
\usepackage{latexsym}

\usepackage[T1]{fontenc}

\usepackage[utf8]{inputenc}

\usepackage{inconsolata}
\usepackage{microtype}
\usepackage{graphicx}
\usepackage{amsmath}
\usepackage{amssymb}
\usepackage{booktabs}

\usepackage{multirow}
\usepackage{utfsym}
\usepackage{appendix}
\usepackage{dutchcal}
\usepackage{bm}
\usepackage{booktabs,makecell,multirow}
\usepackage{xcolor}

\title{Knowledge Generation for Zero-shot Knowledge-based VQA}

\author{Rui Cao \and Jing Jiang \\
  School of Computing and Information Systems \\
  Singapore Management University \\
  \texttt{ruicao.2020@phdcs.smu.edu.sg}, \texttt{jingjiang@smu.edu.sg} \\}

\begin{document}
\maketitle
\begin{abstract}
Previous solutions to knowledge-based visual question answering~(K-VQA) retrieve knowledge from external knowledge bases and use supervised learning to train the K-VQA model.
Recently pre-trained LLMs have been used as both a knowledge source and a zero-shot QA model for K-VQA and demonstrated promising results.
However, these recent methods do not explicitly show the knowledge needed to answer the questions and thus lack interpretability.
Inspired by recent work on knowledge generation from LLMs for text-based QA, in this work we propose and test a similar knowledge-generation-based K-VQA method, which first generates knowledge from an LLM and then incorporates the generated knowledge for K-VQA in a zero-shot manner. We evaluate our method on two K-VQA benchmarks and found that our method performs better than previous zero-shot K-VQA methods and our generated knowledge is generally relevant and helpful.~\footnote{Code available: https://github.com/abril4416/KGen\_VQA}
\end{abstract}

\section{Introduction}
\label{sec:intro}

\begin{figure}[t] 
	\centering
	\includegraphics[width=0.9\linewidth]{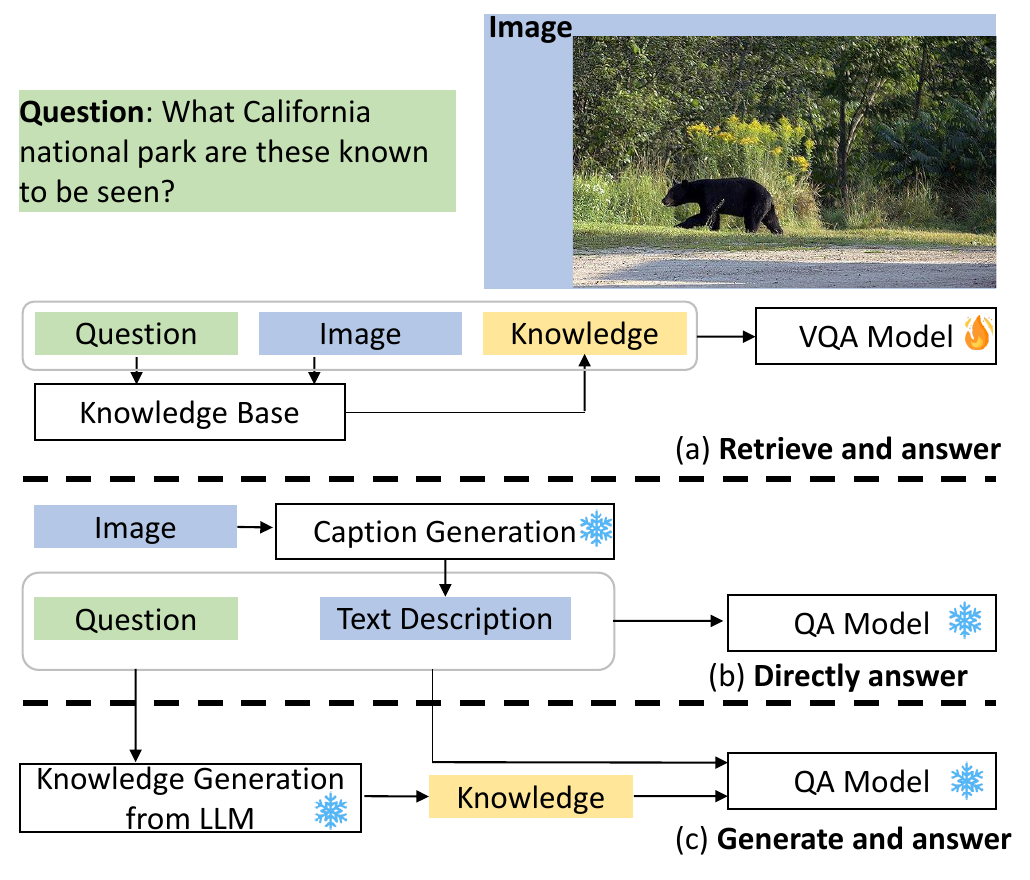} 
	\caption{Three approaches to K-VQA: retrieve and answer, directly answer, and generate and answer. 
 }
	\label{fig:intro-img}
\end{figure}

Knowledge-based VQA~(which we refer to as K-VQA in this paper) is a special visual question answering~(VQA) task where, in addition to an image, external knowledge is needed to answer the given question.
For instance, to answer the question in Figure~\ref{fig:intro-img}, background knowledge about national parks in California is needed.


Early methods for K-VQA follow a \textit{retrieve and answer} paradigm~(Figure~\ref{fig:intro-img}(a)), 
which first retrieves knowledge from external knowledge sources 
as additional input and then trains a VQA model through supervised learning~\cite{DBLP:journals/pami/WangWSDH18,DBLP:conf/eccv/NarasimhanS18,DBLP:conf/nips/NarasimhanLS18,DBLP:conf/mm/Li0020}.
This paradigm requires both a suitable external knowledge base and a large amount of K-VQA training data, which may not be practical for real applications when either of these resources is not available.
Recently, with the fast advances of LLMs that have demonstrated remarkable zero-shot transfer capabilities,
several studies applied LLMs for K-VQA under zero-shot or few-shot settings, leveraging both the extensive knowledge implicitly contained in LLMs and their built-in question answering capability~\cite{DBLP:conf/aaai/YangGW0L0W22,DBLP:journals/corr/abs-2211-09699,DBLP:journals/corr/abs-2212-10846,DBLP:journals/corr/abs-2301-12597,DBLP:conf/nips/AlayracDLMBHLMM22}.
Typically, these methods first convert an image to text descriptions (i.e., captions) and then feed the captions and the question into an LLM 
to directly obtain the answer, as illustrated as the \textit{directly answer} paradigm in Figure~\ref{fig:intro-img}(b).

However, none of these zero-shot or few-shot methods \textit{explicitly} states the knowledge needed to answer a question. 
As we know, answering K-VQA questions usually requires external knowledge not seen in the image. Even if the external knowledge is implicitly contained in the LLM used for QA, it is not immediately clear whether and how the LLM can use the relevant knowledge to answer a K-VQA question through the \textit{directly answer} paradigm. On the other hand, recent work has shown that for text-based QA that requires multi-step reasoning, explicitly generating relevant knowledge and including it as additional input improves QA performance~\cite{DBLP:conf/acl/0010LLWWBCH22, DBLP:journals/corr/abs-2209-10063}.
We suspect that this is also the case for K-VQA. Furthermore, explicitly generated knowledge improves the explainability of the system.
Another limitation of previous zero-shot and few-shot K-VQA methods is that some of them rely on task-specific training such as the training of a question-specific caption generation model in PromptCap~\cite{DBLP:journals/corr/abs-2211-09699}, which still requires significant amount of training data.

In this paper, we attempt to address these limitations of previous work. 
Inspired by \citet{DBLP:conf/acl/0010LLWWBCH22}, which uses an LLM to generate explicit knowledge statements to facilitate text-based commonsense QA, we propose a similar zero-shot K-VQA method that uses an LLM (specifically GPT-3) to \textit{explicitly} generate potentially useful knowledge statements to facilitate K-VQA, as illustrated in Figure~\ref{fig:intro-img}(c).
In addition to having explicit knowledge statements, our method is also free from any additional training.
To improve the diversity and coverage of the generated knowledge, we further borrow the self-supervised knowledge diversification strategy from~\cite{DBLP:journals/corr/abs-2209-10063}.
We call our method \textsc{KGenVQA}.
To the best of our knowledge, we are the first to test the \textit{generate and answer} approach on K-VQA.

We evaluate \textsc{KGenVQA} on both OK-VQA~\cite{DBLP:conf/cvpr/MarinoRFM19} and A-OKVQA~\cite{DBLP:conf/eccv/SchwenkKCMM22}, 
two benchmark datasets commonly used for K-VQA.
The experiments demonstrate that our generated knowledge statements are effective in improving the K-VQA performance in terms of answer accuracy, when everything else being equal, and our method can outperform SOTA zero-shot K-VQA methods that do not use extra training.
We also measure the usefulness of our generated knowledge
and find that the generated knowledge statements have high quality in terms of grammaticality, relevance, factuality, helpfulness, and diversity, based on manual judgement.
Our findings demonstrate that \emph{generate and answer} is a feasible zero-shot approach to K-VQA with the additional benefit of providing explanations through the explicitly generated knowledge statements.


\section{Related Work}
\label{sec:related}


\paragraph{K-VQA.}
Early K-VQA models were built through standard supervised training, with a large amount of (Image, Question, Answer) triplets as training data~\cite{DBLP:journals/pami/WangWSDH18,DBLP:conf/eccv/NarasimhanS18,DBLP:conf/nips/NarasimhanLS18,DBLP:conf/mm/Li0020}.
Typically, these models retrieve knowledge from an external knowledge source such as ConceptNet or Wikipedia and use the retrieved knowledge to facilitate QA.
In our work, we also use explicit knowledge to facilitate QA, but the knowledge is generated from an LLM instead.


\paragraph{Zero-shot K-VQA.}
Several recent studies utilized LLMs for zero-shot K-VQA~\cite{DBLP:conf/aaai/YangGW0L0W22,DBLP:journals/corr/abs-2211-09699,DBLP:journals/corr/abs-2212-10846,DBLP:journals/corr/abs-2301-12597,DBLP:conf/nips/AlayracDLMBHLMM22}.
Generally, these methods first convert the given image into captions or embeddings compatible with a pre-trained language model.
Then the captions or embeddings are combined with the question as input to the language model for zero-shot QA.
We can categorize 
these methods into two types:
those that need extra training using labeled data other than K-VQA data, and those that directly leverage existing pre-trained models without any further training or fine-tuning. 
Examples of the former category include Frozen~\cite{DBLP:conf/nips/TsimpoukelliMCE21} (which uses image-text pairs to train a projection module)
and BLIP-2~\cite{DBLP:journals/corr/abs-2301-12597} (which 
learns a Q-transformer module to model multimodal interactions).
Examples of the latter category include PICa~\cite{DBLP:conf/aaai/YangGW0L0W22} and PNP-VQA~\cite{DBLP:conf/emnlp/Tiong0LSH22}, which convert the images into captions with an off-the-shelf caption generator.
However, to the best of our knowledge, none of the existing zero-shot K-VQA methods explicitly state the external knowledge used to answer the questions.

\paragraph{Knowledge generation for QA.}
A few recent studies on text-based QA tested the idea of using LLMs to generate either short knowledge statements or long documents before combining them with the questions for zero-shot commonsense QA or open-domain QA~\cite{DBLP:conf/acl/0010LLWWBCH22,DBLP:journals/corr/abs-2210-01296,DBLP:journals/corr/abs-2209-10063}.
They found that by incorporating the generated knowledge in QA, performance can be significantly improved. 
Our work is inspired by these recent studies but we apply the idea to visual QA.
\section{Method}
\label{sec:method}

The high-level idea of our \textsc{KGenVQA} method is to leverage an LLM to generate explicit knowledge statements given an image and a question.
These knowledge statements can then be combined with the image captions and the question to be passed to the same or a different LLM for zero-shot text-based QA.
In this section, we first elaborate how we generate knowledge statements from an LLM using few-shot in-context learning.
We then present how the generated knowledge is integrated into the question answering process.

\begin{figure*}[t] 
	\centering
	\includegraphics[width=0.9\linewidth]{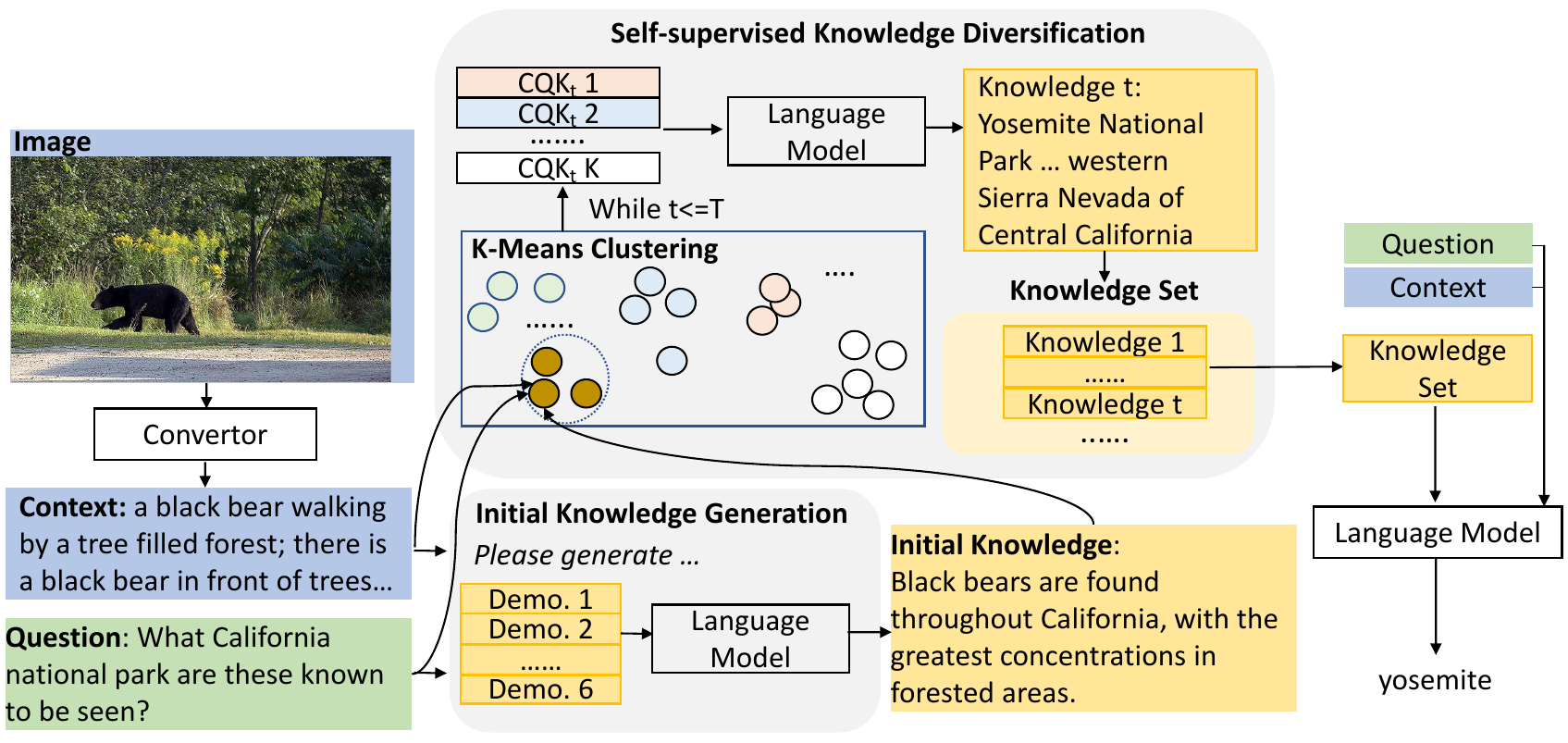} 
	\caption{An overview of the proposed method. We first convert the image into textual descriptions and prompt LLMs with the question and manual demonstrations to obtain the initial knowledge pieces. In the second stage,
 we diversify knowledge by selecting a diverse set of knowledge statements in the first step as demonstrations. Lastly, we incoporate the generated knowledge for QA with a language model.
 }
	\label{fig:arch}
\end{figure*}

\subsection{Knowledge Generation}
\label{sec:knowledge-gen}

Our knowledge generation process consists of two steps: 
An \textit{initial knowledge generation} step, in which we generate a single knowledge statement for each (image, question) pair in the K-VQA test dataset, and a subsequent \textit{self-supervised knowledge diversification} step, in which we sample a diverse set of knowledge statements generated during the first step as in-context demonstrations to perform a second round of knowledge generation, in which we generate multiple knowledge statements per (image, question) pair.
The motivation is that with a diverse set of in-context demonstrations, we expect the LLM to also generate knowledge statements covering different aspects of the same (image, question) pair, which may increase the chance of getting the correct answer.


\paragraph{Caption generation.} In both knowledge generation steps, we regard an LLM (GPT-3 in our experiments) as a knowledge base because the LLM has been trained on a large amount of text covering a wide range of topics.
Previous work has shown that relevant knowledge statements can be generated from an LLM if appropriate text prompts including both the contexts and some demonstrations are used~\cite{DBLP:conf/acl/0010LLWWBCH22}.
However, different from text-based QA, for K-VQA, the context is an image, which cannot be directly used as input to an LLM.
To address this issue, we adopt a simple solution that converts the image into one or more captions, using an off-the-shelf image captioning model.
However, instead of using a general-purpose captioning model, we believe that \textit{question-aware} captions, which focus on describing the parts of the image that are more relevant to the question, can provide better contexts for knowledge generation.
Therefore, we adopt the question-aware caption generation mechanism by \citet{DBLP:conf/emnlp/Tiong0LSH22}, which first highlights image regions that are more relevant to the question and then generates question-aware captions with the attention-weighted image. 
Following the practice of \citet{DBLP:conf/emnlp/Tiong0LSH22}, we use multiple captions because this practice has been shown to be useful for subsequent question answering.
We concatenate the multiple captions into a single sequence of tokens, which we denote as $C$.

\paragraph{Prompt template for knowledge generation.} 
In both the initial knowledge generation step and the knowledge diversification step, to generate a single piece of knowledge, we use the following prompt template:
\textit{Please generate related background knowledge to the question}; \textit{Context:} [$C$]; \textit{Question:} [$Q$]; \textit{Knowledge:}. 
The LLM will complete the prompt above by generating a sentence, which we treat as a knowledge statement. 
In order to better generate the relevant knowledge, we leverage in-context learning by including a few demonstrations, i.e., a few examples each containing a context (which are also image captions), a question, and the expected knowledge statement to be generated.
During the initial knowledge generation step and the knowledge diversification step, we use different kinds of demonstrations.

\paragraph{Initial knowledge generation.} 
During the initial knowledge generation step, we use six manually crafted in-context demonstrations for knowledge generation.
They can be found in Appendix~\ref{sec:manual-prompt}.
During this step, we generate a single knowledge statement for each (image, question) pair in a K-VQA test dataset.



\paragraph{Self-supervised knowledge diversification.}
Previous work showed that proper selection of demonstrations is of vital importance when prompting LLMs~\cite{DBLP:conf/aaai/YangGW0L0W22,DBLP:journals/corr/abs-2212-04037}. 
We suspect that the manually crafted demonstrations may not always be proper examples for all test instances.
Besides, when answering knowledge-intensive questions, oftentimes more than one piece of knowledge may be needed. 
For instance, to answer the question in Figure~\ref{fig:arch}, the knowledge 1) what national parks are in California; 2) among national parks in California, which is famous for black bears.
To generate multiple knowledge statements per question, a straightforward solution is to 
ask the LLM to return multiple pieces of knowledge. 
However, beam search sampling, as mentioned in~\cite{DBLP:conf/iclr/HoltzmanBDFC20,DBLP:conf/aaai/VijayakumarCSSL18}, tends to generate dull and repetitive outputs, and the improved top-$k$ sampling~\cite{DBLP:conf/acl/LewisDF18} can only solve the issue to some extent. 
On the other hand, with different prompts, an LLM may generate diverse outputs~\cite{li2023making}.

Therefore, we 
adopt a self-supervised knowledge diversification strategy by \cite{DBLP:journals/corr/abs-2209-10063} as follows.
Let $\mathcal{K}_\text{init} = \{(C_i, Q_i, K_i)\}_{i=1}^N$ denote the set of (captions, question, knowledge statement) triplets obtained during the initial knowledge generation step, where $K_i$ is the knowledge statement generated for $(C_i, Q_i)$.
We treat each triplet $(C_i, Q_i, K_i)$ as a ``silver''-labeled demonstrating example.
Slightly different from \cite{DBLP:journals/corr/abs-2209-10063}, we hypothesize that if each time we sample a different set of the triplets from $\mathcal{K}_\text{init}$ as demonstrating examples for knowledge generation, and we repeat this $T$  times for a given (image, question) pair $(I, Q)$, then we can obtain $T$ diversified knowledge statements for $(I, Q)$. 
To further ensure that every time the demonstrating examples themselves are diverse, we first use $K$-means clustering to cluster the triplets in $\mathcal{K}_\text{init}$.
Denote these $K$ clusters as $\mathcal{K}_\text{init}^1, \mathcal{K}_\text{init}^2, \ldots, \mathcal{K}_\text{init}^K$.
To generate $T$ final knowledge statements for a given $(I, Q)$ pair during the knowledge diversification step, we repeat the following process $T$ times: (1) we randomly select one triplet from each $\mathcal{K}_\text{init}^k$, except the cluster the given $(I, Q)$ pair belonging to, to form $K-1$ demonstrating examples; (2) we use these $K-1$ demonstrations as in-context examples to generate a knowledge statement for $(I, Q)$, using the prompt template as described earlier.
We call this strategy \textit{self-supervised} knowledge diversification because we do not require any human to annotate diversified demonstrating examples.
We will empirically compare this cluster-based strategy with a random demonstration selection strategy in our experiments.
Details of how $K$-means clustering is done can be found in Appendix~\ref{sec:details-k-means}.

\subsection{Knowledge Integration for K-VQA}

With the final set of $T$ knowledge statements generated for each (image, question) pair, we can combine them with the image captions and the question, and pass them to a pre-trained text-based QA model for answer generation.
In our experiments, we use UnifiedQA~\cite{DBLP:conf/emnlp/KhashabiMKSTCH20}, OPT~\cite{DBLP:journals/corr/abs-2205-01068} and GPT-3~\cite{DBLP:conf/nips/BrownMRSKDNSSAA20}.

\section{Experiments}
\label{sec:exp}

\subsection{Datasets and Evaluation Metrics}
To validate our proposed method, we choose two commonly used K-VQA benchmark datasets, namely, OK-VQA~\cite{DBLP:conf/cvpr/MarinoRFM19} and A-OKVQA~\cite{DBLP:conf/eccv/SchwenkKCMM22}. 
Questions in OK-VQA need outside knowledge beyond the images to answer.
A-OKVQA is an augmented version of OK-VQA that requires additional types of world knowledge.
Because the ground-truth answers of the \textit{test-split} of A-OKVQA are not available, we use its \textit{val-split} for evaluation. 
In the end, the OK-VQA and A-OKVQA datasets we use contain $5,046$ and $1,100$ questions, respectively. 
We report the soft accuracy~\cite{DBLP:conf/cvpr/GoyalKSBP17} on both datasets as there are multiple ground-truth answers for a question. Due to the limit of space, implementation details are provided in Appendix~\ref{sec:exp-settings}.


\subsection{Zero-shot Methods for Comparison}

In this work, we focus on zero-shot K-VQA. 
There are models that need extra training (with labeled data other than K-VQA data).
There are also some few-shot K-VQA methods where the few shots are dynamically selected from a large pool of training examples, which means they still need much training data.
For fair comparison, we do not include these methods because they are not strictly zero-shot. 

Below we briefly review three existing zero-shot K-VQA methods that we compare with:

\noindent\textbf{PICa}~\cite{DBLP:conf/aaai/YangGW0L0W22} converts images into captions with an off-the-shelf caption generator, CLIPCap~\cite{DBLP:journals/corr/abs-2111-09734}. 
The captions are regarded as contexts and fed to GPT-3 together with the question for answer prediction. 

\noindent\textbf{PNP-VQA}~\cite{DBLP:conf/emnlp/Tiong0LSH22} uses improved caption generation by exploiting an image-text matching model~\cite{DBLP:conf/icml/0001LXH22} to highlight image regions related to the question. 
The attended images are then used for caption generation with BLIP~\cite{DBLP:conf/icml/0001LXH22} so that the captions are question-aware. 
We adopt the same caption generation method in PNP-VQA in our method. 
PNP-VQA uses UnifiedQA~\cite{DBLP:conf/emnlp/KhashabiMKSTCH20}, a pre-trained question answering model, in a fusion-in-decoder (FiD) manner~\cite{DBLP:conf/eacl/IzacardG21}, for final answer prediction.

\noindent\textbf{Img2LLM}~\cite{DBLP:journals/corr/abs-2212-10846} follows the caption generation process in PNP-VQA. 
Based on the captions, it generates synthetic QA pairs as demonstrating examples when prompting the LLM for final answers. 
OPT~\cite{DBLP:journals/corr/abs-2205-01068} is used as the LLM for QA.


\begin{table}[t]
  \centering
  \begin{small}
  \begin{tabular}{c|c|c|cc}
    \toprule
   \multicolumn{2}{c|}{\textbf{Model and Size}} & \textbf{Setting} &  \textbf{OK-VQA} &  \textbf{A-OKVQA} \\
   \midrule
    UnifiedQA & \multirow{2}{*}{0.7B} 
     &\textit{w/o} KGen & 32.3 &29.0\\
     & &\textit{w} KGen & 39.7 &31.6\\
     \cmidrule{2-5}
     & \multirow{2}{*}{3B} 
     &\textit{w/o} KGen & 39.6 &35.5\\
     & &\textit{w} KGen & 44.5 &36.5\\
     \cmidrule{2-5}
     & \multirow{2}{*}{11B} 
     &\textit{w/o} KGen & 43.7 &38.9\\
     & &\textit{w} KGen & 45.4 &39.1\\
     \midrule
    OPT & \multirow{2}{*}{6.7B} 
     &\textit{w/o} KGen & 35.2 & 32.4\\
     & &\textit{w} KGen & 39.2& 35.9 \\
     \cmidrule{2-5}
     & \multirow{2}{*}{13B} 
     &\textit{w/o} KGen&37.3 & 35.1 \\
     & &\textit{w} KGen&40.2 &  36.0\\
     \cmidrule{2-5}
     & \multirow{2}{*}{30B} 
     &\textit{w/o} KGen &  37.7&34.4\\
     & &\textit{w} KGen &  42.2&38.1\\
    \bottomrule
\end{tabular}
\end{small}
\caption{Performance comparison between using and not using generated knowledge. KGen refers to knowledge generation.}
\label{tab:gen-effect}
\end{table}


\subsection{Main Results}
\label{sec:gen-effect}


In this section, we empirically evaluate our \textit{generate and answer} approach in two ways: (1) We test the usefulness of the generated knowledge for K-VQA by systematically comparing our K-VQA system with and without knowledge generation. (2) We compare our \emph{generate and answer} method with SOTA zero-shot K-VQA baselines, which do not explicitly generate knowledge. 

\paragraph{The effect of knowledge generation.}

We first conduct systematic experiments to compare the \textit{generate and answer} approach and the \textit{directly answer} approach based on our own implementation. 
To see whether knowledge generation can consistently help K-VQA, we experiment with three different pre-trained QA models: UnifiedQA, OPT, and GPT-3. 
We choose these models because they are used in previous zero-shot K-VQA methods, namely, PNP-VQA, Img2LLM, and PICa, respectively.
When using UnifiedQA, we follow \citet{DBLP:conf/emnlp/Tiong0LSH22} and adopt the FiD strategy.
When using OPT, we follow \citet{DBLP:journals/corr/abs-2212-10846} and add synthetic QA pairs as demonstrations.\footnote{We used the authors' code for synthetic QA pair generation. However, due to different implementation details and the different numbers of synthetic QA pairs used, the performance of our re-implemented Img2LLM base model differs from the reported performance.}

We first show the results of UnifiedQA and OPT on both datasets in Table~\ref{tab:gen-effect}. 
We can see that under all settings (with different QA models and different model sizes), using the generated knowledge consistently improved the final accuracy of the answers. 
For GPT-3, due to the API cost, we only use the first $500$ questions in OK-VQA for performance comparison.
We find that on these 500 test examples, the answer accuracy increased from $\mathbf{27.4}$ to $\mathbf{34.1}$, after adding generated knowledge.
\begin{table}[t]
\centering
\begin{small}
  \begin{tabular}{l|c}
    \toprule
    \textbf{LLM}&\textbf{Num. Kn.} \\
    \midrule
    w/o Gen. Kn. &39.6 \\
    LLaMA$_{7\text{B}}$ &42.1 \\
    LLaMA$_{13\text{B}}$ &42.5 \\
    GPT-3 & 44.5\\
    \bottomrule
\end{tabular}
\end{small}
\caption{Results on OK-VQA when using generated knowledge from different models. \textit{w/o Gen. Kn.} denotes without using any generate knowledge. The text-based QA model is UnifiedQA$_{3\text{B}}$.}
  \label{tab:llama-kb}
\end{table}

Recently, a few open-source LLMs such as LLaMA~\cite{DBLP:journals/corr/abs-2302-13971} have demonstrated comparable performance to GPT-3. We have also considered LLaMA as an alternative choice to GPT-3 for knowledge generation. We incorporate the generated knowledge into UnifiedQA$_{3\text{B}}$ for answer prediction. The results from using LLaMA generated knowledge are provided  in Table~\ref{tab:llama-kb}. According to the results, we can conclude that incorporating generated knowledge from open-source LLMs also benefits K-VQA. By increasing the size of the LLMs, the generated knowledge can more effectively facilitate the model to arrive at the final prediction.
In summary, the results demonstrate that the \textit{generate and answer} approach consistently outperforms the \textit{directly answer} approach on both benchmark datasets under different settings. 

Although our main focus is the zero-shot setting, we also experiment with the few-shot setting, and we find that there is consistent improvement of the \textit{generate and answer} approach over the \textit{directly answer} approach in the few-shot setting, indicating the generalization of our method to few-shot settings. Details of our few-shot experiments can be found in Appendix~\ref{sec:few-shot-setting}.


\begin{table}[t]
\centering
\begin{small}
  \begin{tabular}{lc}
    \toprule
    \textbf{Model} & \textbf{Accuracy} \\
    \midrule
    \multicolumn{2}{c}{\textit{Previous Zero-shot Models \textbf{without} Extra Training}}\\ 
    PICa$_{\text{zero}, \text{175B}}$&17.7\\ 
    PNP-VQA$_{0.7\text{B}}$&27.1\\ 
    PNP-VQA$_{3\text{B}}$&34.1\\ 
    PNP-VQA$_{11\text{B}}$&35.9\\ 
    Img2LLM$_{6.7\text{B}}$&38.2\\
    Img2LLM$_{13\text{B}}$&39.9\\
    Img2LLM$_{30\text{B}}$&41.8\\
    \midrule
    \multicolumn{2}{c}{\textit{KGenVQA (Ours)}}\\ 
    UnifiedQA$_{3\text{B}}$ & 44.5\\
    UnifiedQA$_{11\text{B}}$ & $\mathbf{45.4}$\\
    OPT$_{30\text{B}}$ & 42.2\\
    \midrule
    \midrule
    \multicolumn{2}{c}{\textit{Zero-shot Models \textbf{with} Extra Training}}\\ 
    BLIP-2(OPT)$_{6.7\text{B}}$ &36.4\\ 
    BLIP-2(FlanT5$_\text{XL}$)$_{3\text{B}}$ &  40.7\\ 
    BLIP-2(FlanT5$_\text{XXL}$)$_{11\text{B}}$ &45.9\\ 
    Flamingo$_{\text{3B}}$ & 41.2\\ 
    Flamingo$_{\text{9B}}$ & 44.7\\ 
    \midrule
     \multicolumn{2}{c}{\textit{Few-shot Models} (n=1)}\\ 
    PICa$_{\text{few}, \text{175B}}$ &40.8\\
    PromptCap$_\text{175B}$& $\mathbf{48.7}$\\ 
    \bottomrule
\end{tabular}
\end{small}
\caption{Comparison with SOTA on OK-VQA.
}
  \label{tab:results-okvqa}
\end{table}

\begin{table}[t]
  \small
  \centering
  \begin{tabular}{lc}
    \toprule
    \textbf{Model} & \textbf{Accuracy}\\
    \midrule
     \multicolumn{2}{c}{\textit{Zero-shot Models \textbf{without} Extra Training}}\\ 
     Img2LLM$_\text{6.7B}$ &32.3\\
     Img2LLM$_\text{13B}$ &33.3\\
     Img2LLM$_\text{30B}$  &36.9\\
     \midrule
     \multicolumn{2}{c}{\textit{KGenVQA (Ours)}}\\ 
     UnifiedQA$_{3\text{B}}$ &36.5 \\
    UnifiedQA$_{11\text{B}}$ &$\mathbf{39.1}$ \\
    OPT$_{30\text{B}}$ & 38.1\\
    \midrule
    \midrule
     \multicolumn{2}{c}{\textit{Few-shot Models} (n=10, 32 respectively)}\\ 
    PICa$_\text{few}$ &18.1 \\
    PromptCap$_\text{175B}$ &$\mathbf{56.3}$ \\
    \bottomrule
\end{tabular}
\caption{Comparison with SOTA on A-OKVQA.}
\label{tab:results-aokvqa}
\end{table}


\paragraph{Comparison with SOTA.}
Next, we compare our method with the state-of-the-art models. 
Because we focus on zero-shot K-VQA without extra training, we only compare with previous models of this nature. 
The comparison is shown in the top half of Table~\ref{tab:results-okvqa} for OK-VQA and top half of Table~\ref{tab:results-aokvqa} for A-OKVQA.
We can observe the following from the tables: (1) On both datasets, our \emph{KGenVQA} performs better than the zero-shot baselines when model sizes are comparable. For example, on OK-VQA, our UnifiedQA 3B surpasses all previous zero-shot baselines, i.e., baselines shown in the first block of Table~\ref{tab:results-okvqa}. On A-OKVQA, our UnifiedQA 3B only loses out to Img2LLM 30B, but this is expected because of huge difference of model size. Our method with larger model sizes (i.e., our UnifiedQA 11B and OPT 30B) outperform all zero-shot baselines without extra training.

We also show those zero-shot models with extra training~(e.g., BLIP-2~\cite{DBLP:journals/corr/abs-2301-12597}, Flamingo~\cite{DBLP:conf/nips/AlayracDLMBHLMM22}) and few-shot learning models~(e.g., PICa$_\text{few}$~\cite{DBLP:conf/aaai/YangGW0L0W22} and PromptCap~\cite{DBLP:journals/corr/abs-2211-09699}).
It is worth noting that strictly speaking, PICa$_\text{few}$~\cite{DBLP:conf/aaai/YangGW0L0W22} and PromptCap~\cite{DBLP:journals/corr/abs-2211-09699} do not use the same set of few shot examples~(i.e., is not few-shot learning in the traditional sense) because these two methods dynamically sample demonstrating examples from the whole K-VQA training set for each test example.
Because of their benefits from either extra training or access to the entire training set, we place these models in a different category, at the bottom half of Table~\ref{tab:results-okvqa} and Table~\ref{tab:results-aokvqa}.
Compared with these
models, we can see that our KGenVQA models still surpass some models with extra training, such as
BLIP-2~(FlanT5$_{\text{XL}}$) and the powerful 3B Flamingo, and achieve comparable results with 9B Flamingo, demonstrating the effectiveness of our model compared with state-of-the-art models.
Even comparing with few-shot models,  we observe that our best performance is higher than PICa$_\text{few}$~\cite{DBLP:conf/aaai/YangGW0L0W22} and is comparable to PromptCap$_{175\text{B}}$. 

It may be worth noting that on OK-VQA, PICa$_\text{zero}$ performs poorly probably because it uses a single image caption. 
In order to make a fair comparison with PICa$_\text{zero}$, 
we provide results of our method with a single image caption and without image descriptions (i.e., with generated knowledge only) in Appendix~\ref{sec:fair-comp-pica}.
The results show steady improvements (about $\mathbf{16}$ percentage points in terms of absolute accuracy) on OK-VQA.


\subsection{Ablation Studies}
\begin{table}[t]
\centering
\begin{small}
  \begin{tabular}{l|cc}
    \toprule
    \textbf{Case}&\textbf{Num. Kn.} & \textbf{OK-VQA} \\
    \midrule
    Manual &1 &35.9  \\
    Random
    & 10 & 41.8\\
    CoT
    &1 & 37.5 \\
    KGen &10 &$\mathbf{44.8}$ \\
    \bottomrule
\end{tabular}
\end{small}
\caption{Comparison of different knowledge generation methods on OK-VQA. ``Num. Kn.'' is the number of knowledge statements used.}
  \label{tab:cluster-kb}
\end{table}

\noindent\textbf{Knowledge generation method.} We first compare our cluster-based knowledge diversification strategy with (1) using the manual prompt generated knowledge, i.e., a single piece of knowledge (Manual); 
(2) randomly sampling $K-1$ single knowledge statement, instead of sampling from different clusters, from the initially generated knowledge statements, $\mathcal{K}_\text{init}$ for knowledge diversification in the second stage (Random).
Besides, we consider the idea of Chain-of-Thoughts (CoT)~\cite{DBLP:conf/nips/Wei0SBIXCLZ22}, which generates explanations before the answer generation. 
In K-VQA, the needed knowledge can also be regarded as a kind of explanations. Therefore, we test the widely used CoT for knowledge generation, which is an alternative to our cluster-based knowledge generation approach. 
We re-use the six manual demonstrations as mentioned in Section~\ref{sec:method} and manually add answers to the questions (i.e., each demonstration consists of contexts of image descriptions, a question, a piece of related knowledge and an answer). 
Together with these demonstrations, we prompt GPT-3~\cite{DBLP:conf/nips/BrownMRSKDNSSAA20} to first generate the relevant knowledge and then the answer (CoT). 
Due to the cost of calling GPT APIs, we only apply CoT to a subset questions on OK-VQA (200 questions). 
We show model performance, based on UnifiedQA$_{3\text{B}}$, with different ways of knowledge generation and show results in Table~\ref{tab:cluster-kb}. 
We have a few observations: 
(1) using initial generated knowledge with demonstrations offers improvements but no better than KGen. This may be that fixed manual demonstrations fail to generate diverse knowledge. For a fair comparison, we also consider using a single piece of knowledge from KGen, which achieves $\mathbf{38.8}$, indicating the need of diverse prompts in knowledge generation.
(2) Comparing using random selection and cluster-based selection in the self-supervised knowledge diversification stage, we find that using the cluster-based method clearly outperforms random selection, which may not generate diverse knowledge. 
Overall, the cluster-based knowledge generation method is better than the other methods for knowledge generation in term of K-VQA performance;
(3) When we compare the CoT knowledge generation with cluster-based knowledge generation, the second method significantly wins CoT in terms the benefit to K-VQA, probably because the cluster-based method has higher chances of facilitating answer generation with diverse knowledge; 
Besides, we also compare the direct CoT-generated answers from GPT-3 with answers generated when prompting GPT-3 for QA incorporating our generated knowledge. 
Our generated knowledge results in an accuracy of 32.0 while CoT-generated knowledge leads to 29.3.

\begin{table}[t]
\centering
\begin{small}
  \begin{tabular}{l|cc}
    \toprule
    \textbf{QA Model}&\textbf{Num.} & \textbf{OK-VQA}\\
    \midrule
    \multirow{4}{*}{UnifiedQA (FiD)$_{3\text{B}}$}&
    0 & 39.6\\
    &5 & $\mathbf{44.5}$\\
    &10 & $\mathbf{44.5}$\\
    &20 &42.7\\
    \midrule
    \multirow{4}{*}{OPT$_{13\text{B}}$}&
    0 & 37.3\\
    &5 & $\mathbf{40.2}$\\
    &10 & 37.2\\
    &20 &37.2\\
    \midrule
    \multirow{4}{*}{GPT-3}&
    0 & 27.4\\
    &5 & $\mathbf{34.1}$\\
    & 10 & 32.4\\
    &20 &31.7\\
    \bottomrule
\end{tabular}
\end{small}
\caption{Performances with different numbers of knowledge statements. 
}
  \label{tab:num-kb}
\end{table}

\noindent\textbf{Number of knowledge statements.}
Next, we test how the number of knowledge statements affects the performance, using UnifiedQA$_{3\text{B}}$ (FiD), OPT$_{13\text{B}}$ and GPT-3. 
Due to the API costs, we choose OK-VQA as the experiment dataset for this ablation study. 
For GPT-3 as the QA model, we test the performance on the first $500$ questions. 
The results are reported in Table~\ref{tab:num-kb}. 
Intuitively, we observe improvements after adding more generated knowledge at first and then decrement of performance. 
This is probably because adding too many pieces of knowledge may potentially add noisy or redundant knowledge, which harms the performance. 
Besides, we notice that decoder-only models have smaller optimal number of knowledge statements than encoder-decoder FiD model. 
This is probably because decoder-only models (i.e., OPT and GPT-3) may have difficulty in understanding the long concatenated sentence while FiD is specifically designed for comprehension of multiple documents.

\begin{table}[t]
\centering
\begin{small}
  \begin{tabular}{l|cccc}
    \toprule
    \textbf{Case}&\textbf{Gram.} & \textbf{Rel.} & \textbf{Fact.} & \textbf{Help.}\\
    \midrule
    Ours$_\text{max}$ &100.0& 100.0&  96.3& 90.0\\
    Ours$_\text{avg}$ &99.0 & 100.0&94.5 & 67.0\\
    \midrule
\end{tabular}
\end{small}
\caption{Evaluation of our generated knowledge in terms of four evaluation metrics.}
  \label{tab:human-eval}
\end{table}

\subsection{Evaluation of the Generated Knowledge}
In this section, we conduct human evaluation to exam the quality of the generated knowledge. We follow \citet{DBLP:conf/acl/0010LLWWBCH22} and sample $40$ cases from OK-VQA dataset where the correctness of the answers would be changed (i.e., either from correct to wrong or wrong to correct) after adding the generated knowledge. 
For each instance, we sample $5$ knowledge statements for evaluation.
We ask two annotators to check the quality of the generated knowledge in terms of the evaluation metrics below. 
To ensure objectiveness, annotators will not know whether the predictions are changed to become correct or wrong. 

\noindent\textbf{Evaluation metrics.} Following \citet{DBLP:conf/acl/0010LLWWBCH22,DBLP:conf/emnlp/ShwartzWBBC20}, we take four metrics for evaluating generated knowledge: 1) \textit{Grammatically}: whether it is grammatical  2) \textit{Relevance}: whether it is related to answering the question and the image; 3) \textit{Factuality}: whether it is factual; 4) \textit{Helpfulness}: whether it is helpful so that it directly leads to the correct answers or provides indirect but supportive information of the correct answers. 
For \textit{helpfulness}, we adopt three categories of evaluation: helpful (i.e., provides direct or indirect supportive information to correct answers), harmful (i.e., negates correct answers or support incorrect answers) or neutral (neither helpful or harmful). 
Besides the previously used metrics, we also consider \textit{Diversity} as the fifth evaluation criteria, indicating the coverage of generated knowledge.
Details about the definitions can be found in Appendix~\ref{sec:human-eval-details} and the examples we provide to annotators regarding the four evaluation metrics are included in the supplementary materials. 

\noindent\textbf{Results.} The average agreement from two annotators over four evaluation metrics is $0.67$, in terms of \textit{Fleiss Kappa} $\kappa$~\cite{landis1977measurement}. 
It indicates substantial agreement among annotators. 
For each criterion, we report the average score over two annotators. 
We consider two evaluation settings for generated knowledge: 1) \textit{average}: taking the average scores over five pieces or knowledge; 2) \textit{max}: take the maximum score over scores of five knowledge. 
The results are provided in Table~\ref{tab:human-eval}. 
According to the results, most knowledge is grammatical, relevant to questions and factual. 
One interesting thing is that the generated knowledge may be relevant to questions but harmful for final answers, as the average score in term of \textit{helpfulness} is only around 70.
From the comparison with \textit{average} and \textit{max} scores of human evaluation, we further verify the need of knowledge diversification, which can raise the chance of generating helpful knowledge, as indicated by the maximum score of \textit{helpfulness}, which means how likely the generated knowledge will lead to the correct answer.
For diversity, we compare the five pieces knowledge generated by cluster-based selection against random selection. The average diversity of cluster-based select is $\mathbf{3.4}$, while $\mathbf{2.5}$ for random selection. It shows cluster-base selection results in more diverse knowledge, which is more likely to cover information for answering questions. It is in consistency with results in Table~\ref{tab:cluster-kb}.

\subsection{Case Study}
\label{sec:error-analysis}
To better understand the advantage of our method, we compare our method with the baseline, UnifiedQA$_{3\text{B}}$ (FiD), without generated knowledge. We analyze the first $20$ cases, without cherry picking, where our method answers correctly while the baseline gives wrong predictions. Among the $20$ error cases of the baseline, $85\%$ are due to the lack of external knowledge, highlighting the advantage of our method.
Due to the limitation of space, we provide the examples in Appendix~\ref{sec:case-comparison}.

Besides, we conduct error analysis to better understand the limitations of our method. We conduct an empirical analysis for the error cases by manual checking $40$ error cases from UnifiedQA$_{3\text{B}}$ (FiD) after adding generated knowledge. Among all error cases, we observe $20\%$ are due to the undesired knowledge. Due to limitation of space, we provide visualization of the error cases in Appendix~\ref{sec:error-analysis}. The main cause of generating misleading knowledge comes from the inaccurate image descriptions which lack details for LLMs for knowledge generation. It implies with the development of better image description generation tools, our method can be potentially improved.

\section{Conclusions}

In this work, we propose to generate relevant knowledge from LLMs for zero-shot K-VQA. 
We evaluate the effectiveness of the generated knowledge by experimenting with different pre-trained QA models of varying model sizes on two K-VQA benchmarks. 
The experiment results show that the generated knowledge improves K-VQA performance, and our method can outperform SOTA zero-shot K-VQA methods.
We further conduct human evaluation to validate the quality of the generated knowledge. 
The results demonstrate that the generated knowledge statements are relevant and helpful to questions in K-VQA.
\section{Limitations}

In this paper, we adopt GPT-3.5 as the LLM to generate several pieces of knowledge for one question. 
However, the generated knowledge may be redundant in some cases, which introduces noise to the final answer prediction process. 
Therefore, in the future, we need to investigate how to filter out redundant knowledge.
Besides, in this work we only consider inserting the generated knowledge into a text-QA model when converting K-VQA into a text-based QA problem. 
A future direction is to design and insert generated knowledge into pre-trained vision-language models (PT-VLMs)~(e.g., BLIP-2~\cite{DBLP:journals/corr/abs-2301-12597}), because the conversion from images to texts may leave out crucial details, but PT-VLMs can take images as inputs without losing any potentially important visual information from the images. 

\section*{Acknowledgement}

This research was supported by the Ministry of Education, Singapore, under its Academic Research Fund Tier 2 (Grant No.: T2EP20222-0047, Project ID: MOE-000440-00).  Any opinions, findings and conclusions or recommendations expressed in this material are those of the authors and do not reflect the views of the Ministry of Education, Singapore.

\bibliography{anthology}
\bibliographystyle{acl_natbib}

\clearpage
\appendix
\section{Details of K-Means Clustering}
\label{sec:details-k-means}
To divide testing instances into different clusters, we first convert each context-question-knowledge triplet into vector representations. Specifically, the context, question and the initial piece of knowledge will be concatenated and the textBERT~\cite{devlin2018bert} to encode the concatenated sentence. Based on the encoded textual representation, we used the \textit{K-Means} clustering to divide all instances into $K$ clusters. Given an instance waiting for knowledge generation, which belongs to the cluster $k$, instances from other clusters will serve as demonstrations. In other words, we randomly select one demonstration from each cluster except the $k$-th cluster so that there are $K-1$ demonstrations for the testing example. The set of demonstrations we denote as \texttt{PSEUDO DEMO}. Then we prompt LLMs again with the self-supervised demonstrations with an input. We will iteratively conduct the process mentioned above T times where at the $t$-th time step we obtain a piece of knowledge $\mathcal{k}_t$ and finally we have $T$ knowledge pieces.
\begin{table}[t]
  \centering
  \begin{small}
  \begin{tabular}{c|c|c|c|c}
    \toprule
   \multicolumn{2}{c|}{\textbf{Model and Size}} &\textbf{\# shots}& \textbf{Setting} &  \textbf{OK-VQA}  \\
   \midrule
    OPT & \multirow{1}{*}{13B} 
    &32 &\textit{w/o} KGen & 36.1 \\
     & &32&\textit{w} KGen & 39.6 \\
     \cmidrule{2-5}
     & \multirow{1}{*}{30B} 
     &16&\textit{w/o} KGen & 36.7 \\
     & &16&\textit{w} KGen & 43.8 \\
    \bottomrule
\end{tabular}
\end{small}
\caption{Performance comparison between using and not using generated knowledge in the few-shot setting on OK-VQA dataset. KGen refers to knowledge generation.}
\label{tab:few-shot-okvqa}
\end{table}

\begin{table}[t]
\centering
\begin{small}
  \begin{tabular}{lc}
    \toprule
    \textbf{Model} & \textbf{Model Size} \\
    \multicolumn{2}{c}{\textit{Zero-shot Models \textbf{without} Extra Training}}\\ 
    PICa$_\text{zero}$&175B\\ 
    PNP-VQA&1.2B, 3.4B, 11.8B \\ 
    Img2LLM& 6.7B, 13B, 30B, 66B, 175B\\
    \midrule
    \midrule
    \multicolumn{2}{c}{\textit{Zero-shot Models with Extra Training}}\\ 
    VL-T5$_\text{no-vqa}$&269M\\  
    Frozen &7.1B\\ 
    VLKD$_\text{ViT-L/14}$&832M\\
    FewVLM& 785M\\ 
    BLIP-2(OPT$_{6.7\text{B}}$) &7.8B\\ 
    BLIP-2(FlanT5$_\text{XL}$) &4.1B  \\ 
    BLIP-2(FlanT5$_\text{XXL}$)&12.1B\\ 
    Flamingo &3B, 9B, 80B \\ 
    \midrule
     \multicolumn{2}{c}{\textit{Few-shot Models}}\\ 
     ClipCap$\to$Cap.$\to$GPT & 175B\\
    ClipCap$\to$Ratl.$\to$GPT & 175B\\
     PICa$_\text{few}$ &175B\\
    PromptCap& 175B\\ 
    \midrule
\end{tabular}
\end{small}
\caption{Summarizing of models for K-VQA.}
  \label{tab:model-info}
\end{table}

\begin{table}[t]
\centering
\begin{small}
  \begin{tabular}{lc}
    \toprule
    \textbf{Model} & \textbf{Acc.} \\
    \midrule
    \multicolumn{2}{c}{\textit{Zero-shot Models \textbf{without} Extra Training}}\\ 
    PICa$_{\text{zero},175B}$&17.7\\ 
    PNP-VQA$_{0.7\text{B}}$&27.1\\ 
    PNP-VQA$_{3\text{B}}$&34.1\\ 
    PNP-VQA$_{11\text{B}}$&35.9\\ 
    Img2LLM$_{6.7\text{B}}$&38.2\\
    Img2LLM$_{13\text{B}}$&39.9\\
    Img2LLM$_{30\text{B}}$&41.8\\
    Img2LLM$_{66\text{B}}$&43.2\\
    Img2LLM$_{175\text{B}}$&45.6\\
    \midrule
    \multicolumn{2}{c}{\textit{Zero-shot Models with Extra Training}}\\ 
    VL-T5$_\text{no-vqa}$&5.8\\  
    Frozen &5.9\\ 
    VLKD$_\text{ViT-L/14}$&13.3\\
    FewVLM& 16.5\\ 
    BLIP-2(OPT)$_{6.7\text{B}}$ &36.4\\ 
    BLIP-2(FlanT5$_\text{XL}$)$_{3\text{B}}$ &  40.7\\ 
    BLIP-2(FlanT5$_\text{XXL}$)$_{3\text{B}}$ &45.9\\ 
    Flamingo$_{3\text{B}}$ & 41.2\\ 
    Flamingo$_{9\text{B}}$ & 44.7\\ 
    Flamingo$_{80\text{B}}$ & 50.6\\ 
    \midrule
     \multicolumn{2}{c}{\textit{Few-shot Models}}\\ 
    PICa$_{\text{few},175B}$ (n=1) &40.8\\
    PromptCap$_{175B}$ (n=1) & 48.7\\ 
    \midrule
\end{tabular}
\end{small}
\caption{Model performancee on OK-VQA dataset. For models with different model sizes, we show the model size with subscripts.}
  \label{tab:full-results-okvqa}
\end{table}

\begin{table}[t]
\centering
\begin{small}
  \begin{tabular}{lc}
    \toprule
    \textbf{Model} & \textbf{Acc.} \\
    \midrule
    \multicolumn{2}{c}{\textit{Zero-shot Models \textbf{without} Extra Training}}\\ 
    Img2LLM$_{6.7\text{B}}$&33.3\\
    Img2LLM$_{13\text{B}}$&33.3\\
    Img2LLM$_{30\text{B}}$&36.9\\
    Img2LLM$_{66\text{B}}$&38.7\\
    Img2LLM$_{175\text{B}}$&42.9\\
    \midrule
     \multicolumn{2}{c}{\textit{Few-shot Models}}\\ 
    ClipCap$\to$Cap$\to$GPT$_{175\text{B}}$ (n=10)&16.6\\
    ClipCap$\to$Rel$\to$GPT$_{175\text{B}}$&18.1\\
    PromptCap$_{175B}$ (n=32)& 56.3\\ 
    \midrule
\end{tabular}
\end{small}
\caption{Model performancee on A-OKVQA dataset. For models with different model sizes, we show the model size with subscripts. }
  \label{tab:full-results-aokvqa}
\end{table}

\begin{table}[!ht]
\small
  \begin{tabular}{|c|p{2.7cm}|p{2.7cm}| }
    \hline
    \textbf{Img.} & \begin{minipage}[b]{0.37\columnwidth}
		\centering
		\raisebox{-.5\height}{\includegraphics[width=\linewidth]{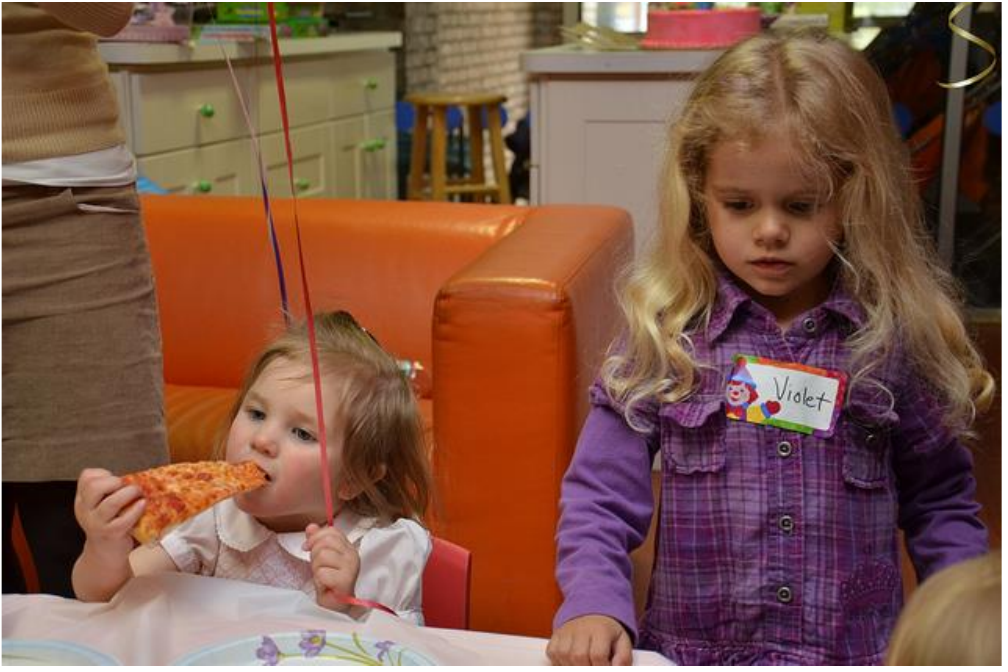}}
	\end{minipage} &
    \begin{minipage}[b]{0.34\columnwidth}
		\centering
		\raisebox{-.5\height}{\includegraphics[width=\linewidth]{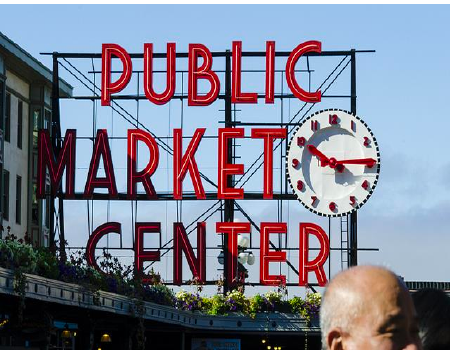}}
	\end{minipage} \\\hline
     \textbf{Ques.} &Which type of leather is used for making the sofa set shown in this picture? &  Where in the world is this located?\\\hline
    \textbf{GT.} & cow, fake, fine grain, suede & seattle, san francisco, seattle usa, boston massachusetts \\\hline
    \textbf{Pred.} &black leather & czech republic \\\hline
    \textbf{Cap.} & two child a pizza pizza three people child up pizza. a young girl and a young girl with pizza as food.
a young girl eating pizza while sitting in a booth
&a sign outside of a market market sign on a clear day.
the sign shows market square, with a lot of people, and a large clock.
a group of people outside of a building showing a clock.\\\hline
    \textbf{Kn.}
    &The sofa set shown in this picture is likely made of faux leather, which is a synthetic material made to look and feel like real leather.
    & This market square is located in the city of Prague, Czech Republic. 
\\\hline

    \end{tabular}
    \centering
    \caption{Visualization of error cases. GT. is for ground-truth annotation, Pred. is for predictions from models, Cap. is for the image captions and Kn. is for generated knowledge.}
  \label{tab:case-error}
\end{table}

\begin{table*}[!ht]
\centering
  \begin{tabular}{|p{2cm}|p{4cm}|p{4cm}|p{4cm}| }
    \hline
    \textbf{Image} & \begin{minipage}[b]{0.5\columnwidth}
		\centering
		\raisebox{-.5\height}{\includegraphics[width=\linewidth]{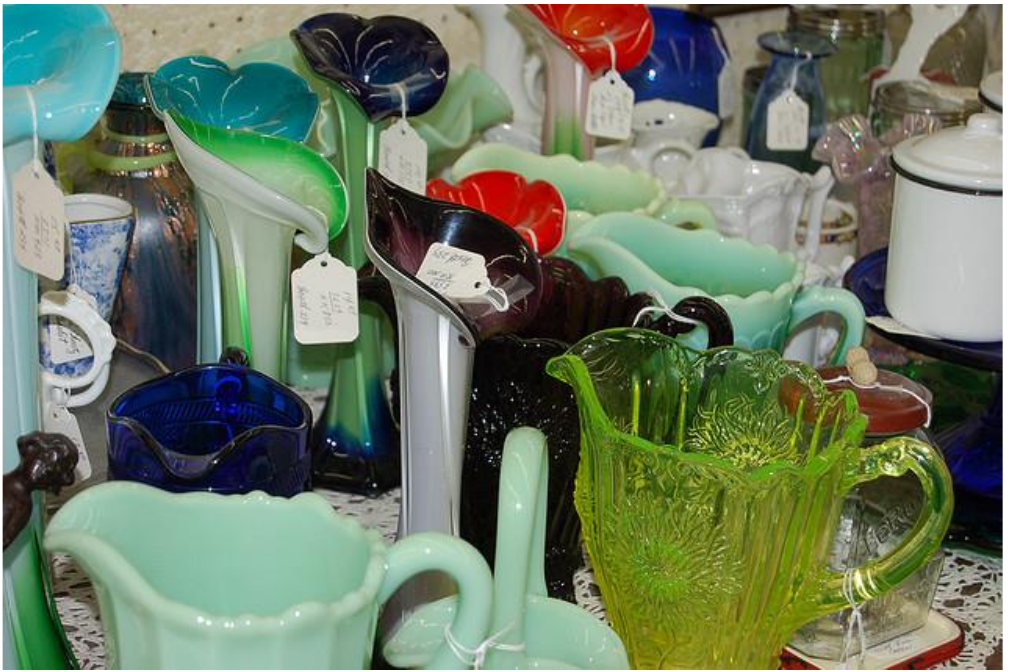}}
	\end{minipage} &
    \begin{minipage}[b]{0.5\columnwidth}
		\centering
		\raisebox{-.5\height}{\includegraphics[width=\linewidth]{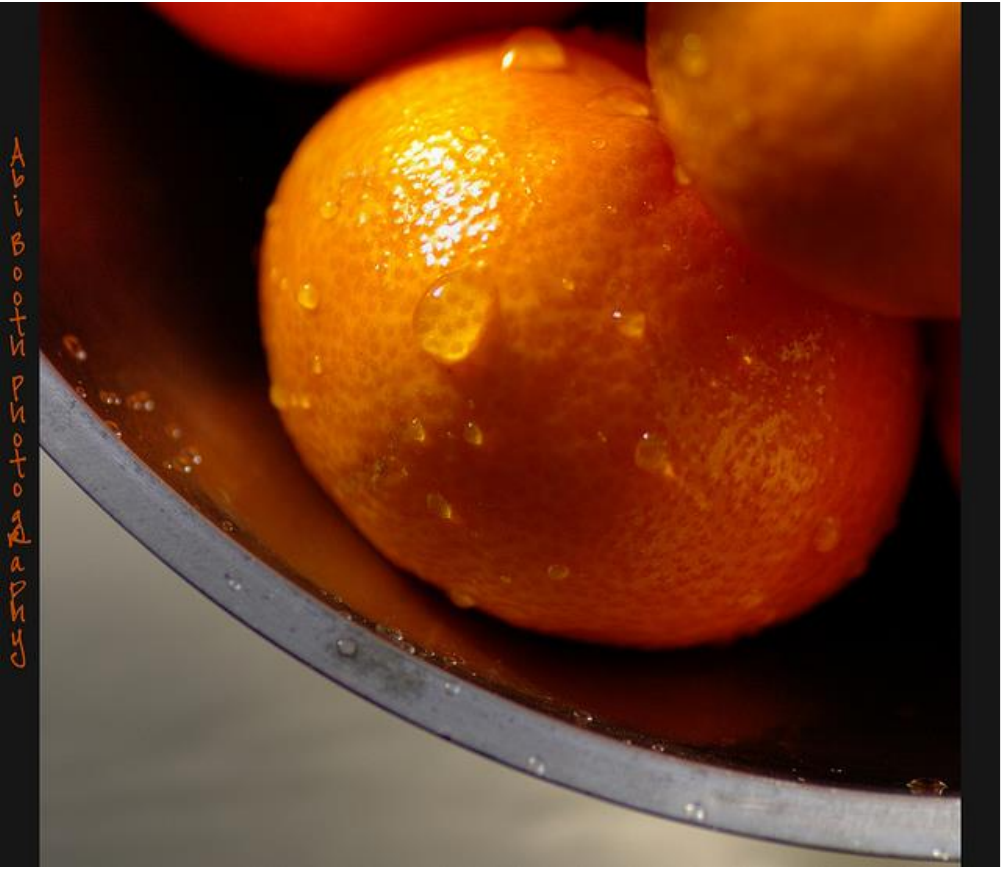}}
	\end{minipage} &
    \begin{minipage}[b]{0.5\columnwidth}
		\centering
		\raisebox{-.5\height}{\includegraphics[width=\linewidth]{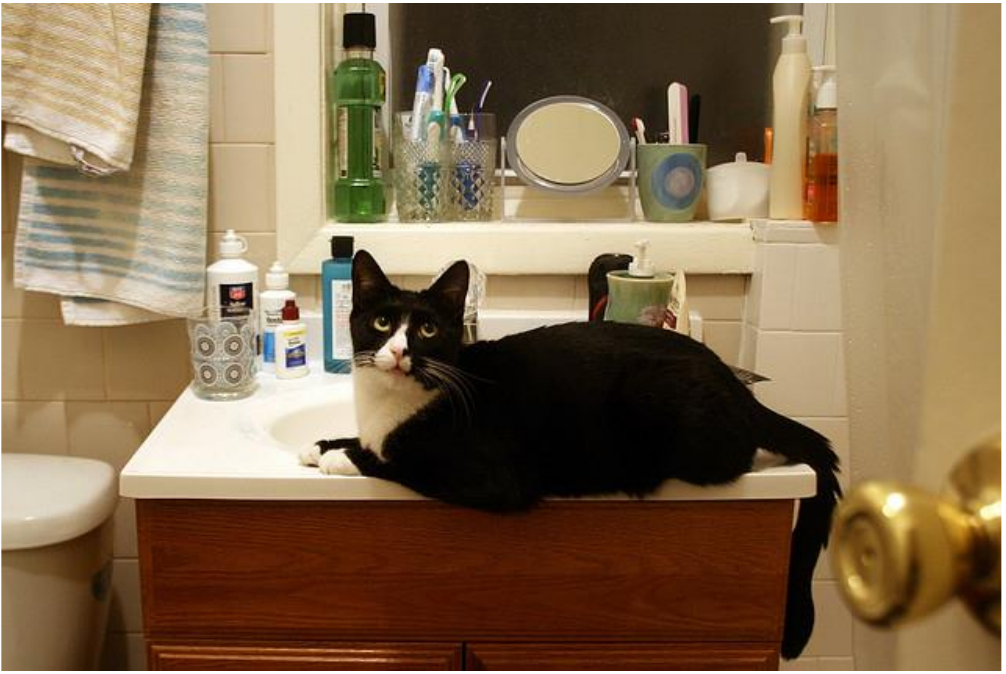}}
	\end{minipage}\\\hline
    \textbf{Question} 
    & What would happen if these items fall to the ground?
    & What sates are these grown in? 
    & Name one famous person whom also has a black and white one of these? \\\hline
    \textbf{Ground Truth}  
    & shatter, they would shatter, break, they would break
    & florida california, california, florida
    & taylor swift, russell brand, hillary clinton, ernest hemingway\\\hline
    \textbf{Base Prediction} 
    & \color{red} nothing
    & \color{red} texas 
    & \color{red} kate winslet\\\hline
    \textbf{Generated Knowledge} 
    &If a glass item falls to the floor, it will break.
    &California and Florida are the leading producers of oranges.
    &Taylor Swift is a famous singer and songwriter who has a black and white cat named Meredith. \\\hline
    \textbf{Our Prediction} 
    &they would break 
    &california 
    &taylor swift\\\hline
    \textbf{Image} & \begin{minipage}[b]{0.5\columnwidth}
		\centering
		\raisebox{-.5\height}{\includegraphics[width=\linewidth]{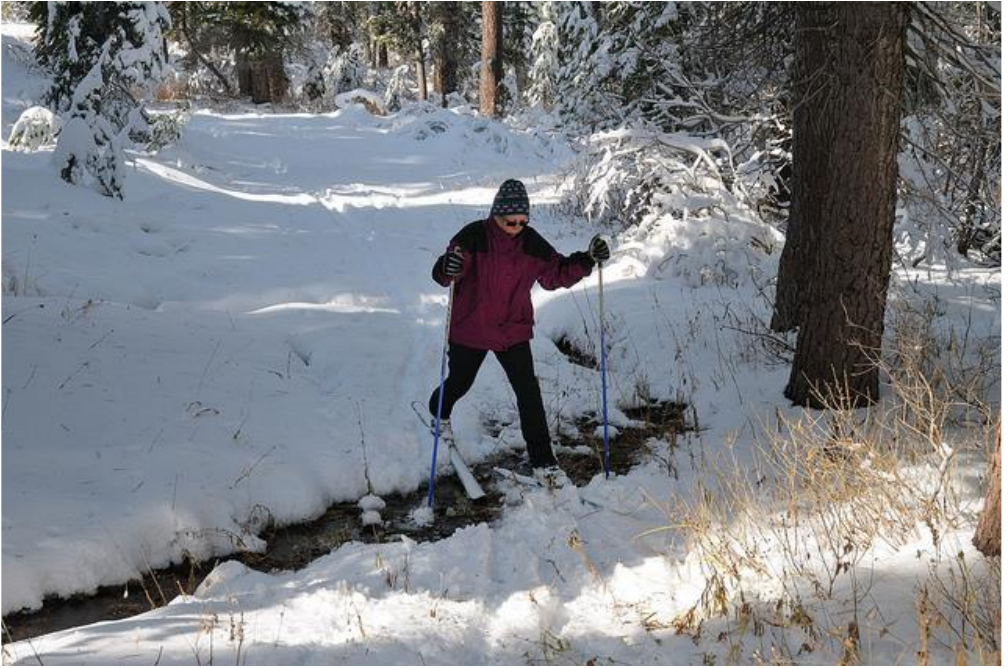}}
	\end{minipage} &
    \begin{minipage}[b]{0.5\columnwidth}
		\centering
		\raisebox{-.5\height}{\includegraphics[width=\linewidth]{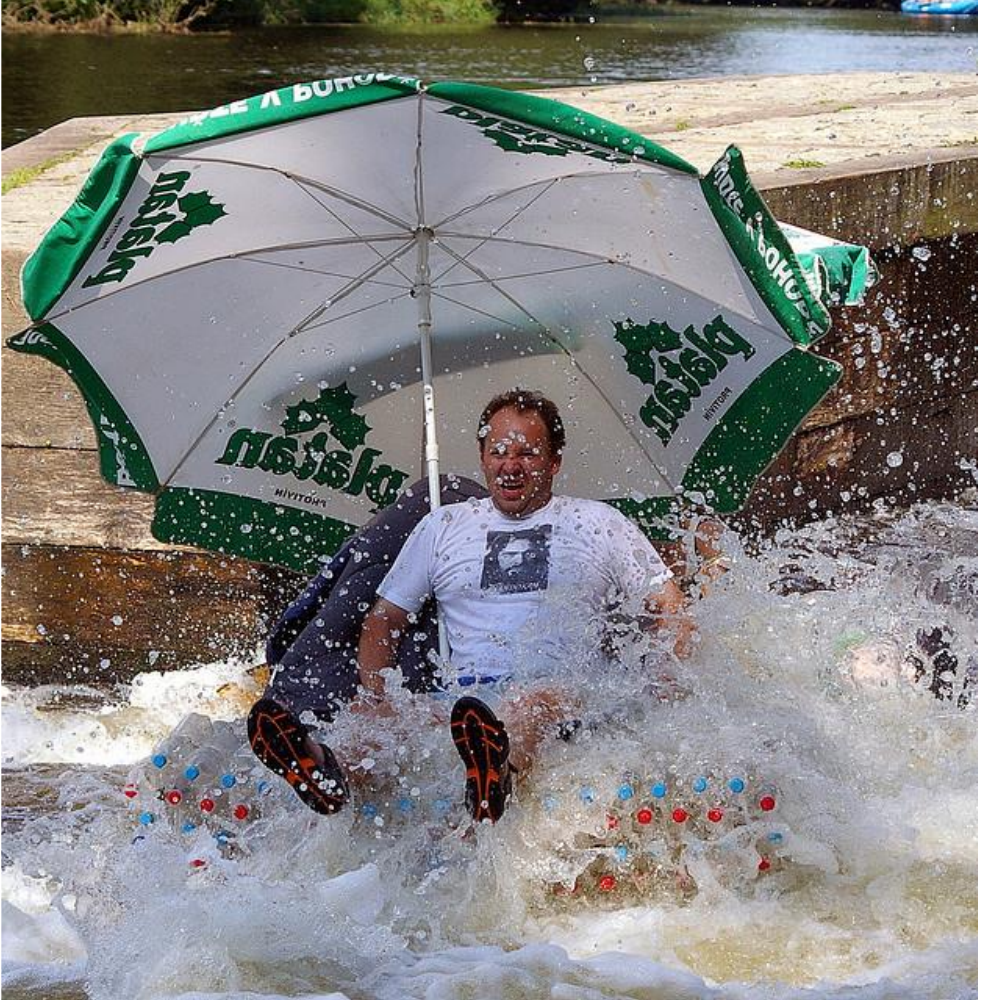}}
	\end{minipage} &
    \begin{minipage}[b]{0.5\columnwidth}
		\centering
		\raisebox{-.5\height}{\includegraphics[width=\linewidth]{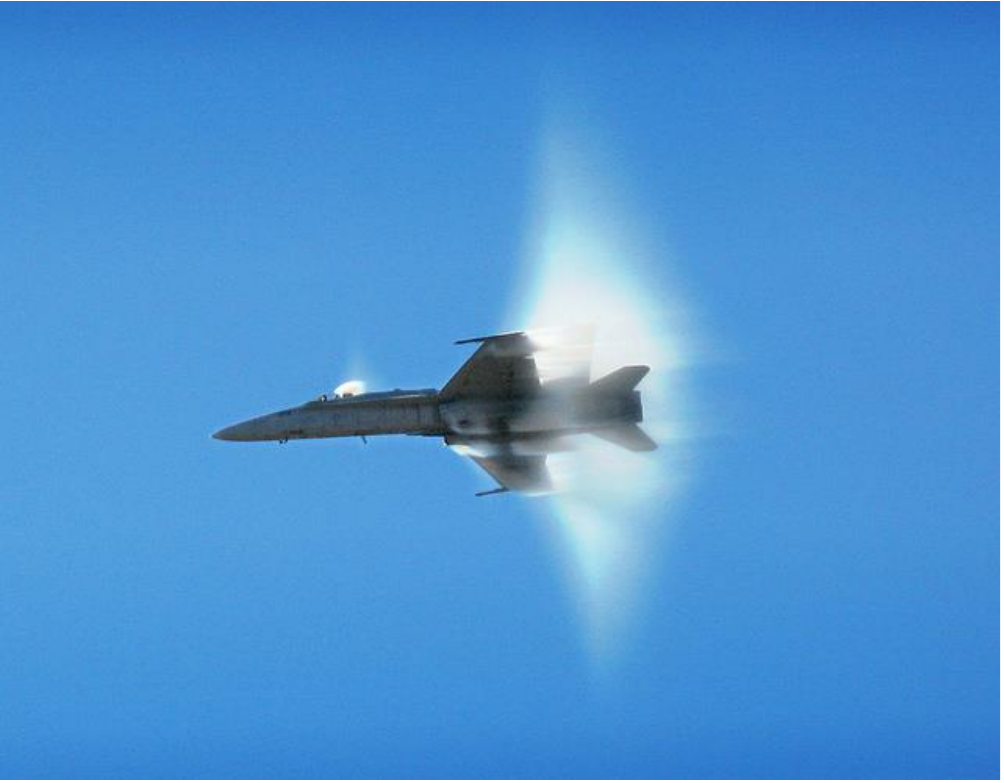}}
	\end{minipage}\\\hline
    \textbf{Question} 
    & If it gets cold enough what will happen to the area being stepped over?
    & What knocked the guy off his chair?
    & What is the white cloud behind the jet called?\\\hline
    \textbf{Ground Truth}  
    & freeze, frozen, it will freeze over, iced
    & wave, water
    & contrail, cloud, supersonic wave\\\hline
    \textbf{Base Prediction} 
    & \color{red} snow 
    & {\color{red} water splash} 
    & \color{red} halo\\\hline
    \textbf{Generated Knowledge} 
    &If it gets cold enough, the area being stepped over will freeze, creating a layer of ice on top of the snow.
    &The waves in the water knocked the man off his chair.
    &The condensation trail, or contrail, is a visible trail of condensed water vapor created by an aircraft engine or wingtip vortices under certain atmospheric conditions. \\\hline
    \textbf{Our Prediction} 
    &frozen
    & wave
    & contrail\\\hline
    \end{tabular}
    \caption{Visualization of error cases of the baseline without generated knowledge, while our method answers correctly with the help with generated knowledge. Wrong predictions are highlighted in {\color{red} red}.}
  \label{tab:case-compare}
\end{table*}

\section{Experiment Settings}
\label{sec:exp-settings}
\noindent\textbf{Experiment Details} 
For knowledge generation, we use GPT-3.5~(\textit{text-davinci-003}\footnote{https://platform.openai.com/docs/models/gpt-3-5}) as our LLM, with a suggested temperature of $0.7$. For the $K$-means clustering in knowledge diversification stage, we set the number of cluster to be $8$ empirically.

For answer prediction, because exact match is adopted for evaluation, we encourage the pre-trained QA model to give short answers. 
For UnifiedQA, we set the length penalty to be -1; for GPT-3.5, we add the following instruction: \textit{Generate answers with as fewer words as possible.} 
After answer prediction, we conduct an answer post-processing step as proposed in~\cite{DBLP:journals/corr/abs-2308-01390}.

We implement our model on NVIDIA Tesla V100 GPUs with 32 GB of dedicated memory. The system ran on CUDA version 11.1. For UnifiedQA, except $11$B version, we implemented with a single GPU. For UnifiedQA $11$B model and OPT model series, we implement with model parallel on four GPUs.

\noindent\textbf{Package Version}  In this experiment, we rely on the PyTorch library, 1.13.1 version. For the implementation of BLIP~\cite{DBLP:conf/icml/0001LXH22} (used for image caption generation), we leverage the LAVIS package from Salesforce~\footnote{https://github.com/salesforce/LAVIS/tree/main/lavis} (version 1.0.2), for OPT~\cite{DBLP:journals/corr/abs-2205-01068} and UnifiedQA model~\cite{DBLP:conf/emnlp/KhashabiMKSTCH20} we use the transformers package from Huggingface~\footnote{https://huggingface.co/} (version 4.29.2), and for GPT-3.5 model, we leverage the OpenAI API~\footnote{https://platform.openai.com/overview}.

\noindent\textbf{Model Size:} We show model size in Table~\ref{tab:model-info}. If we one model has different versions of model size, we separate them with comma. 

\section{Few-shot Setting Results}
\label{sec:few-shot-setting}
We provide the results for our method in the few-shot setting on OK-VQA in the section. Specifically, we leverage the OPT model~\cite{DBLP:journals/corr/abs-2205-01068} as the final QA model and give a few demonstrations. Each demonstration consists of a question, an image description as the context, an answer and optional related knowledge (in the \textit{w} KGen setting).
The results are shown in Table~\ref{tab:few-shot-okvqa}. According to the results, we observe consistent improvements after adding generated knowledge, indicating our method can generalize to the few-shot setting as well.

\section{Fair Comparison with PICa$_{\text{zero,175B}}$}
\label{sec:fair-comp-pica}
Considering PICa$_{\text{zero,175B}}$ leverages only a single piece of image description while our method uses multiple captions, following~\cite{DBLP:conf/emnlp/Tiong0LSH22}, improvements may potentially come from more detailed image descriptions. To ablate the impact from image description side, we use a single caption as the image description, similar to PICa$_{\text{zero,175B}}$. It achieves $\mathbf{33.8}$ on OK-VQA, with about $\mathbf{16}$ absolute accuracy improvements over PICa$_{\text{zero,175B}}$. Further more, we used only the generated knowledge as inputs to text-based QA models (UnifiedQA$_{3\text{B}}$). It achieves $\mathbf{33.5}$ on OK-VQA, highlighting that generated knowledge itself contains information for question answering.

\section{Model Performance}
\label{sec:full-exp-results}
We only provide models in a fair comparison in Section~\ref{sec:gen-effect}. In this part, we provide performance of models on K-VQA including zero-shot K-VQA models without extra training but have larger model sizes, zero-shot K-VQA models with extra training and few-shot K-VQA models. The results on OKVQA and A-OKVQA are shown in Table~\ref{tab:full-results-okvqa} and Table~\ref{tab:full-results-aokvqa} respectively. 

\section{Error Cases}
\label{sec:error-appendix}
In this section, we provide visualization of two error cases of which the generated knowledge is inadequate. The reason of generating the harmful knowledge is because of inaccurate image captions. A potential way of improving our method is to improve the quality of image descriptions. 

\section{Comparison with the Baseline without Knowledge}
\label{sec:case-comparison}
In this section, we provide visualization of error cases of the baseline model without knowledge and compare with our method. The visualized examples are shown in Table~\ref{tab:case-compare}. Noted, we do not perform cherry-picking. The visualized cases are the first six error cases of the baseline model on OK-VQA while being correctly addressed by our method. To keep the table tidy, we only present one piece of generated table. According to the visualization, we observe our generated knowledge largely benefit addressing these questions in need of external knowledge.

\begin{table*}[t]
  \small
  \centering
  \begin{tabular}{l|p{12cm}}
    \hline
    \textbf{Num.} & \textbf{Content}\\
    \midrule
    1&Context:The company in the image is Monsanto. There are two men selling products. The logo behind two men is Monsanto.
Question:What does company in the image own?
Knowledge:Monsanto is a multinational agrochemical and agricultural biotechnology corporation. It is one of the world’s leading producers of roundup, a glyphosate herbicide.\\
\midrule
    2&Context:The red vegetable is tomato. There is a sandwich with tomato and lettuce. There is a sandwich on the table.
Question:Where can this red vegetable be found?
Knowledge:tomatoes are usually planted in gardens.
\\\midrule
    3&Context:The man is playing tennis. The man is holding a tennis racket. A man is in a competition of tennis.
Question:What English city is famous for a tournament for the sport this man is playing? 
Knowledge:The Wimbledon Championships is the oldest tennis tournament in the world.\\\midrule
    4&Context:a plate with ham, tomatoes, meat, and sliced peppers on top of it. breakfast and bacon eggs scrambled toast. a breakfast sandwich, tomatoes, bacon, and eggs
Question:what food in the photo has a lot of c vitamin? 
Knowledge:Tomatoes and tomato products are rich sources of folate, vitamin C, and potassium. Eggs contain decent amounts of vitamin D, vitamin E, vitamin B6, calcium and zinc. Bacon provides a good amount of B vitamins.\\\midrule
    5&Context:a man sitting in front of a laptop computer smiling and posing for the camera. a man wearing glasses sitting in front of a laptop. a man in glasses and glasses at a desk with laptop.
Question:what purpose do the glasses the man is wearing serve? 
Knowledge:Glasses are typically used for vision correction, such as with reading glasses and glasses used for nearsightedness.\\\midrule
    6&Context:a bedroom with a bed, wall paper and lamp. a bed with storage underneath it in a room. a bed in a small room with pillows and box drawers.
Question:what was the largest size of that platform that we have? 
Knowledge:Single size is 91 cm x 190 cm. Super single size is 107 cm x 190 cm. Queen size is 152 cm x 190 cm. King size is 182 cm x 190 cm.\\
    \midrule
\end{tabular}
\caption{Contents of manual prompts.}
\label{tab:manual-prompt-exp}
\end{table*}

\section{Manual Prompts}
\label{sec:manual-prompt}
Here we provide a full list of six manual prompts in Table~\ref{tab:manual-prompt-exp}. Before the demonstrations, we also add an instruction: \textit{Please generate related background knowledge to the question:}  in the front. Knowledge are collected from searching with Google.

\section{Details for Human Evaluation}
\label{sec:human-eval-details}
In this part, we provide more details about human evaluation about the knowledge quality. We invite two annotators for evaluation of $40$ questions with five pieces of generated knowledge. Firstly, they will be given an instruction, indecating the definition of the K-VQA task, an example of the K-VQA task and the goal of the evaluation. Next, we describe what information (i.e., question, ground-truth answer, generated knowledge, and image) will be provided to them and the denotations of the information. Thirdly, we elaborate the definitions of four metrics. For the metrics of \textit{Relevance}, \textit{Factuality} and \textit{Helpfulness}, besides definitions, we provide a few concrete examples in texts to make it easier for understanding. The definifions and examples are provided in Table~\ref{tab:eval-define}. For the full information of the annotated knowledge, please refer to the Supplementary file.
\begin{table*}[t]
  \small
  \centering
  \begin{tabular}{l|p{3cm}|p{8cm}}
    \hline
    \textbf{Attributes} & \textbf{Definition}& $\textbf{Example}$\\
    \midrule
    Grammaticality &Whether the knowledge statement is grammatical (e.g., whether a complete and fluent sentence; whether human can understand the sentence).
    & None
    \\ \midrule
    Relevance &Whether a knowledge statement is relevant to the given question. A statement is relevant if it covers the same topic as the question or contains a salient concept that is the same as or similar to the one in the question (provided indirect but related information). 
    & [Image]: a bedroom with a bed\newline
 [Question]: what was the largest size of that platform that we have?\newline
[Knowledge]: Single size is 91 cm x 190 cm. Super single size is 107 cm x 190 cm. Queen size is 152 cm x 190 cm. King size is 182 cm x 190 cm.\newline
[Judge]:Relevant. Because the information is related to the topic on bed size.

    \\ \midrule
    Factuality &Whether a knowledge statement is (mostly) factually correct or not. If there are exceptions or corner cases, it can still be considered factual if they are rare or unlikely. 
    & [Image]: a triangle in the image
[Question]: what shape is the object in the image?\newline
[Knowledge]: A rectangle is a shape with two equal sides\newline
[Judge]: Not factual, because a rectangle has four sides\newline

[Image]: a limousine; a car\newline
[Question]: how many doors does the vehicle in the image have?\newline
[Knowledge]: A limousine has four doors.\newline
[Judge]: Factual.\newline

[Image]: a human being\newline
[Question]: how many fingers does this creature have?\newline
[Knowledge]: A human hand has four fingers and a thumb.\newline
[Judge]: Factual, despite that there are exceptions – people with disabilities may have less or more fingers.

    \\ \midrule
    Helpfulness &Whether a knowledge statement is (mostly) factually correct or not. If there are exceptions or corner cases, it can still be considered factual if they are rare or unlikely. 
    & [Image]: a subway in the image\newline
[Question]: How often you take this transportation back and forth to work per week?\newline
[Knowledge]: You take the subway back and forth to work five days a week\newline
[Judge]: Helpful. Because the statement directly supports the answer.\newline

[Image]: a spider\newline
[Question]: how many legs does the animal in the image have?\newline
[Knowledge]: Arachnids have eight legs\newline
[Judge]: Helpful. Although the statement does not directly refer to spiders, together with the fact that "spiders are a kind of arachnids" it completes a reasoning chain in deriving the answer.\newline

[Image]: two persons are playing chess\newline
[Question]: what are the results of the game?\newline
[Knowledge]: A game of chess has two outcomes\newline
[Judge]: Harmful. Since the statement supports answering "two outcomes" instead of "three outcomes".\newline

[Image]: a person in the white background.\newline
[Question]: How many chromosomes does the creature have?\newline
[Knowledge]: human beings are mammals.\newline
[Judge]: Neutral. The knowledge does not provide information in favor or contrast of answering the question.

    \\ \midrule
\end{tabular}
\caption{Definitions and examples for evaluation metrics.}
\label{tab:eval-define}
\end{table*}

\end{document}